\documentclass[10pt,journal,twocolumn]{IEEEtran}    

\IEEEoverridecommandlockouts

\usepackage{graphicx}
\usepackage{color}
\usepackage{srcltx}

\newcommand{\R}[1]{\mathrm{#1}}
\newcommand{\B}[1]{\mathbf{#1}}
\newcommand{\AfterTheoremEnvironment}{\vspace{-1em} \hfill $\blacksquare$ }

\usepackage{graphicx,subfigure} 
\usepackage{epsfig}             
\usepackage{mathptmx}           
\usepackage{times}              
\usepackage{amsmath}            
\usepackage{amssymb}            

\newtheorem{lemma}{Lemma}
\newtheorem{example}[lemma]{Example}
\newtheorem{remark}[lemma]{Remark}

\newtheorem{proposition}[lemma]{Proposition}
\newtheorem{assumption}[lemma]{Assumption}

\begin{document}

\title{
The connection between Bayesian estimation of a Gaussian random field
and RKHS
}
\author{
	Aleksandr Y. Aravkin, 
	Bradley M. Bell, 
	James V. Burke and
	Gianluigi Pillonetto
	\thanks{Aleksandr Y. Aravkin (saravkin@us.ibm.com) is with the 
	IBM T.J. Watson Research Center, Yorktown Heights, NY 10598}
	\thanks{Bradley M. Bell (bradbell@uw.edu) is with the 
	Applied Physics Laboratory \&
	Institute for Health Metrics and Evaluation,
	University of Washington, Seattle WA, USA}
	\thanks{James V. Burke  (burke@math.washington.edu) is with the 
	Department of Mathematics, University of Washington, Seattle, WA, USA}
     \thanks{G. Pillonetto (giapi@dei.unipd.it) is with the Department 
	of Information Engineering, University of Padova, Padova, Italy}
	\thanks{This research has been partially supported by the 
	European Community's Seventh Framework Programme [FP7/2007-2013] under
	agreement n. FP7-ICT-223866-FeedNetBack, under
	grant agreement n257462 HYCON2 Network of excellence
	and by the FIRB project entitled ``Learning meets time".}
}

\maketitle \thispagestyle{empty} \pagestyle{empty}

\begin{abstract}
Reconstruction of a function from noisy data 
is often formulated as a regularized optimization problem 
over an infinite-dimensional reproducing kernel Hilbert space (RKHS). 
The solution 
describes the observed data and has a small RKHS norm. 
When the data fit is measured using a quadratic loss, this 
estimator has a known statistical interpretation.
Given the noisy measurements,
the RKHS estimate represents the posterior mean (minimum variance estimate)
of a Gaussian random field with covariance proportional 
to the kernel associated with the RKHS.
In this paper, we provide a statistical interpretation 
when more general losses are used,
such as absolute value, Vapnik or Huber.
Specifically, for any finite set of sampling locations 
(including where the data were collected),
the MAP estimate for the signal samples is given by the 
RKHS estimate evaluated at these locations.

This connection establishes a firm statistical foundation for
several stochastic approaches used
to estimate unknown regularization parameters. 
To illustrate this, we develop a numerical scheme that implements
a Bayesian estimator with an absolute value loss. 
This estimator is used to learn a function
from measurements contaminated by outliers.
\end{abstract}


\section{Introduction}
Minimizing a regularized functional with respect to a reproducing kernel
Hilbert space (RKHS) $\mathcal{H}$
is a popular approach to
reconstruct a function $F: \mathcal{X} \rightarrow \B{R}$ from noisy data;
e.g. see
\cite{Aronszajn,Scholkopf01b,Wahba1990,PoggioGirosi}. 
To be specific,
regularization in $\mathcal{H}$ estimates $F$ using
$\hat{F}$ defined by
\begin{equation}
\label{DefineFhat}
\hat{F} 
=
\arg \min_{ F \in \mathcal{H} } 
	\left( \sum_{i=1}^N V_i [ y_i - F( x_i ) ] 
	+ 
	\gamma \| F \|_{\mathcal{H}}^2 \right) \; , 
\end{equation}
where $\gamma \in \B{R}^+$ is the regularization parameter,
$\mathcal{X}$ is a set (finite or infinite),
$x_i \in \mathcal{X}$ is the location where $y_i \in \B{R}$ is measured,
$V_i : \B{R} \rightarrow \B{R}^+$ is the loss function for $y_i$, and
$\|\cdot\|_{\mathcal{H}}$ is the RKHS norm induced 
by the positive definite reproducing kernel 
$K: \mathcal{X} \times \mathcal{X} \rightarrow \B{R}$, see \cite{Aronszajn}.
Here $y_i$ is the ith element of a vector $y$.
Note however that $x_i$ is the ith measurement location 
(not the ith element of a vector $x$), and 
$V_i$ is the loss function corresponding to the ith residual. 

One of the important features of the above approach is that, even if
the dimension of $\mathcal{H}$ is infinite, the solution belongs
to a finite-dimensional subspace. In fact, under mild assumptions on the loss,
according to the representer theorem \cite{Wahba1998,Scholkopf01},
$\hat{F}$ in \eqref{DefineFhat} is the sum of kernel sections
$K_i : \mathcal{X} \rightarrow \B{R}$ defined by $K_i(x) = K(x_i, x)$.
To be specific,
\begin{equation}
\label{RepresenterEquation}
\hat{F}(\cdot) = \sum_{i=1}^N \hat{c}_i K_i(\cdot) \; ,
\end{equation}
where $\hat{c}$ is defined by
\begin{equation}
\label{DefineChat}
\hat{c} =
\arg \min_{c \in \B{R}^N} \left( 
	\sum_{i=1}^N V_i \left[
		y_i - \sum_{j=1}^N K( x_i, x_j ) c_j 
	\right]
	+ \gamma c^\R{T} \overline{K} c
\right) \; .
\end{equation}
Here and below, $\overline{K} \in \B{R}^{N \times N}$ 
denotes the {\it{kernel matrix}}, or Gram matrix,
defined by $\overline{K}_{ij} = K(x_i, x_j)$.
When the component loss functions $V_i( \cdot )$ are quadratic, 
the problem in \eqref{DefineFhat} 
admits the structure of a regularization network
\cite{Poggio} and also has a statistical interpretation.
Specifically, suppose that
$F$ is a zero-mean Gaussian random field
with a prior covariance proportional to $K$,
and that $F$ is independent of the white Gaussian measurement noise.
Then, given the measurements, 
for every $x$ the value \( \hat{F}(x) \) 
is the posterior mean, and hence the minimum variance estimate of $F(x)$, 
e.g. see subsection 2.3 of \cite{Anderson:1979}.
This connection, briefly reviewed in Section \ref{MapEstimationSection},
is well known in the literature and was initially studied in 
\cite{Kimeldorf71Bayes}
in the context of spline regression, 
see also \cite{Wahba1990,Girosi95,Rasmussen}.
This connection can be proved using the representer theorem,
which also yields the closed form
solutions of the coefficients $\hat{c}_i$ in \eqref{RepresenterEquation}: 
\begin{equation} 
\label{ComputeChat}
\hat{c} = \left( \overline{K}+\gamma\B{I}_N \right)^{-1}y \; ,
\end{equation}
where $y \in \B{R}^N$ is the vector of measurements $y_i$
and $\B{I}_N$ is the $N \times N$ identity matrix.

A formal statistical model for more general loss functions (e.g.,
the Vapnik $\varepsilon$-insensitive loss
used in support vector regression \cite{Vapnik98,Evgeniou99,Gunter06})
is missing from the literature.
After interpreting the $V_i$ 
as alternative statistical models for the observation noise, 
many papers argue that $\hat{F}$ in \eqref{DefineFhat}
can be viewed as a 
maximum a posteriori (MAP)
estimator assuming the a priori probability density of $F$ is
proportional to $\exp(-\| F \|_\mathcal{H}^2)$, 
e.g. \cite[Section 7]{Evgeniou99}. 
These kinds of statements are informal, since 
in an infinite-dimensional function space the concept of probability
density is not well defined, see e.g.
\cite{Bogachev} for a thorough treatment of Gaussian measures.
{\em The main contribution of this note is to provide
a rigorous statistical model that justifies $\hat{F}$
as an estimate of a Gaussian random field.}

This connection provides a firm statistical foundation
for several stochastic approaches
for estimating unknown regularization parameters. 
Examples of such parameters include $\gamma$ in (\ref{DefineFhat}) and 
possibly other parameters used to specify $K$.
To illustrate, we develop a Bayesian estimator
equipped with the absolute value ($\ell_1$) loss
using the Markov chain Monte Carlo (MCMC) framework \cite{Gilks}. 
The estimator recovers a function
starting from measurements contaminated by outliers, 
and compares favorably with the tuning approach
recently proposed in \cite{Dinuzzo07} where $\gamma$ is determined
using CP-like statistics and the concept of equivalent degrees of freedom.

The structure of the paper is as follows.
In Section \ref{StatisticalModelSection} we formulate the statistical model.
In Section \ref{MapEstimationSection}, 
we review the connection between regularized estimation
in RKHS and estimation in the quadratic case,
and then extend this connection to more general losses. 
Section \ref{RegularizationParameterSection} uses this connection to describe
Bayesian approaches that estimate regularization parameters,
in addition to the unknown function.
A numerical experiment is then reported in 
Section \ref{SimulationExampleSection} to illustrate the theoretical results.
Section \ref{ConclusionSection} contains a summary and conclusion.
The proofs are presented in Section \ref{AppendixSection}.

\section{Statistical Model}
\label{StatisticalModelSection}

Here and below, $\B{E}[\cdot]$ indicates the expectation operator,
and given (column) random vectors $u$ and $v$, we define
\[
\R{cov}[u,v]
=
\B{E} \left[ (u-\B{E}[u]) (v-\B{E}[v])^\R{T} \right] \; .
\]
We assume that the measurements $y_i$ are obtained
by measuring the function $F$ at sampled points $x_i$ in the presence
of additive noise, i.e.
\begin{equation} 
\label{MeasMod}
y_i = F(x_i) + e_i, \quad i = 1, \ldots, n \; ,
\end{equation}
where each $x_i$ is a known sampling location.
We make the following assumptions:

\begin{assumption}
\label{GaussianFunctionAssumption}
We are given a known positive definite autocovariance function 
$K$ on $\mathcal{X} \times \mathcal{X}$ and
a scalar $\lambda > 0$ such that for any sequence of 
points $\{ x_j : j = 1 , \ldots , J \}$,
the vector $f = [ F( x_1 ) , \ldots , F( x_J ) ]$ 
is a Gaussian random variable with mean zero and covariance
given by 
\[
\R{cov} ( f_j , f_k ) = \lambda K ( x_j , x_k ) \; . 
\]
\end{assumption}
\AfterTheoremEnvironment

A random function $F$ that satisfies 
Assumption~\ref{GaussianFunctionAssumption} 
is often referred to as a zero-mean Gaussian random field on $\mathcal{X}$.

\begin{assumption}
\label{GeneralMeasurementAssumption}
We are given a sequence of measurement pairs
$( x_i , y_i ) \in \mathcal{X} \times \B{R}$ 
and corresponding loss functions $V_i$ for $i = 1 , \ldots , N$.
In addition, we are given
a scalar $\sigma > 0$ such that
\begin{equation*}
\B{p} (y | F) 
\propto 
\prod_{i=1}^N \exp \left( 
	-\frac{ V_i[ y_i - F( x_i ) ] }{2 \sigma^2} 
\right) \; . 
\end{equation*}
\end{assumption}
Furthermore, the measurement noise random variables
\( e_i = y_i - F( x_i ) \) are independent of the
the random function $F$.

\AfterTheoremEnvironment

\begin{figure}
{\includegraphics[scale=0.24]{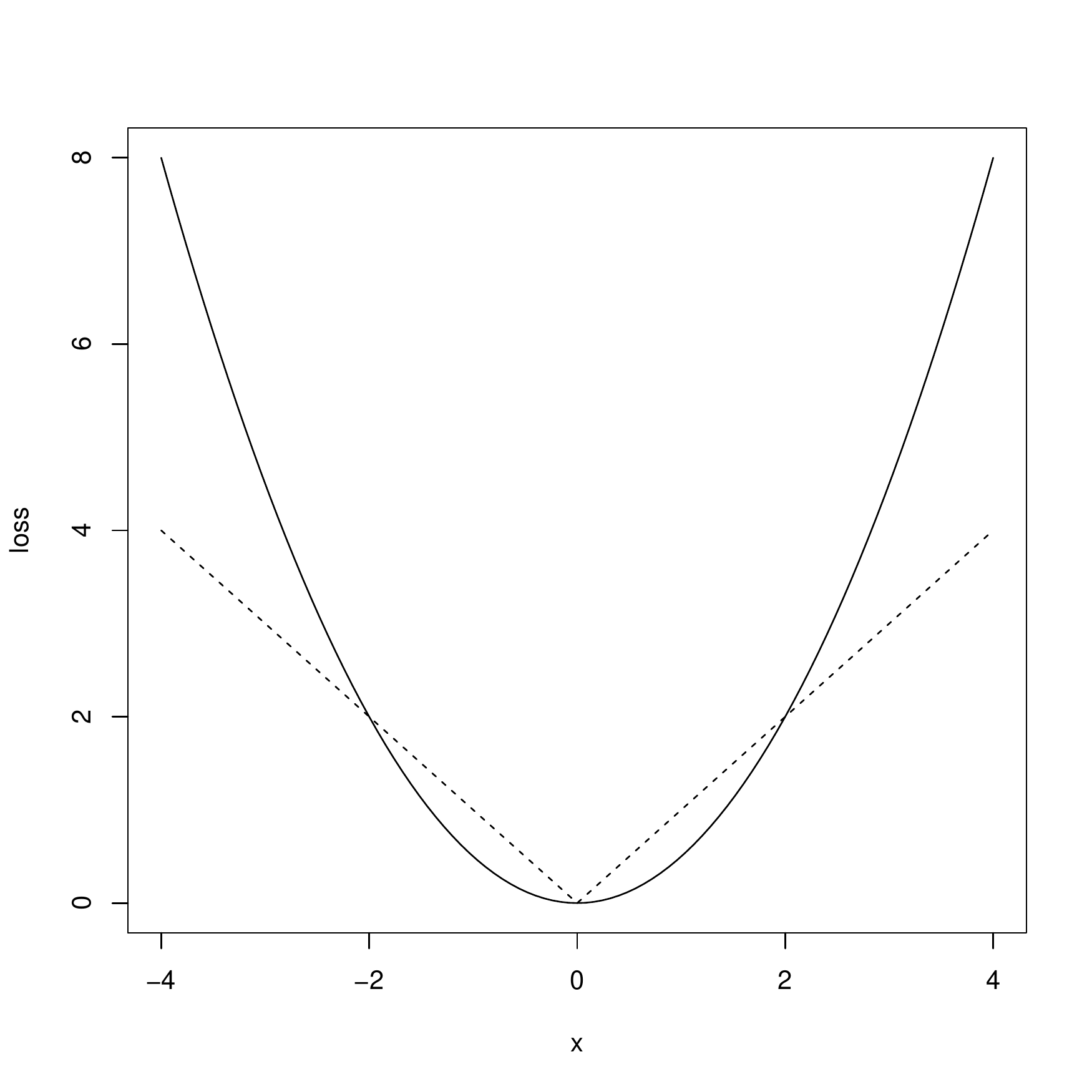}}
{\includegraphics[scale=0.24]{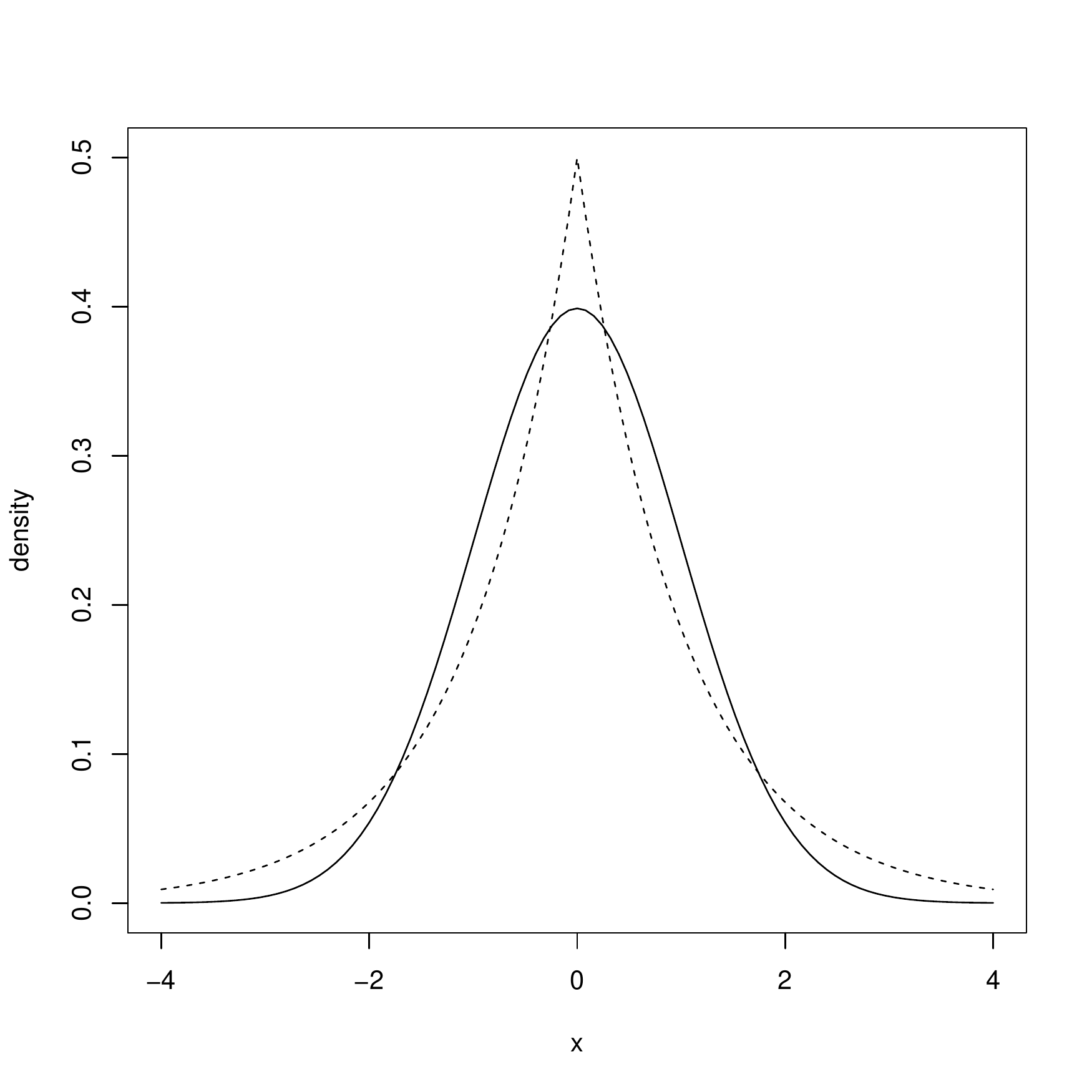}}
\caption{
\label{FigurePenaltyDensity}
Left: quadratic and absolute losses are solid and dashed lines. 
Right: mean zero variance one Gaussian (solid) and Laplace (dashed) densities.
Note that Laplace has heavier tails than the Gaussian,
which explains its robustness properties.
}
\end{figure}

For example, 
$V_i (r) = r^2$ corresponds to Gaussian noise,
while using $V_i(r) =  |r|$ corresponds to Laplacian noise.
These loss functions (and corresponding standardized densities)
are pictured in Figure~\ref{FigurePenaltyDensity}.
The statistical interpretation of an  $\epsilon$-insensitive $V_i$
in terms of Gaussians with mean and variance 
described by suitable random variables
can be found in  \cite{NoisePMG2000}.

\section{Estimation in reproducing kernel Hilbert spaces}
\label{MapEstimationSection}

\subsection{Gaussian measurement noise}

We first consider the case of Gaussian measurement noise; 
i.e., $V_i(r) = r^2$.
This corresponds to modeling the $\{ e_i \}$ as i.i.d. Gaussian random
variables with variance $\sigma^2$. 
In view of the independence of $F$ and $e$, 
it turns out that $F(x)$ and $y$ are
jointly Gaussian for any $x \in \mathcal{X}$. 
Hence, the posterior $\B{p} [ F(x) | y ]$ is also Gaussian.
The mean and variance for this posterior
can be calculated using the following proposition
\cite[Example 3.6]{Anderson:1979}.

\begin{proposition} 
\label{ConditionalProposition}
Suppose $u$ and $v$ are jointly Gaussian random vectors.
Then, $\B{p}( u | v )$ is also Gaussian 
with mean and autocovariance given by
\begin{eqnarray*}
\B{E}  ( u |  v) 
& = & 
\B{E} ( u ) + \R{cov}  ( u, v ) \R{cov}( v, v )^{-1}  [ v  - \B{E} (v) ] \; ,
\\ 
\R{cov} (u , u |v) 
& = & 
\R{cov} ( u, u ) -  \R{cov}  ( u, v )    \R{cov}( v, v )^{-1}  \R{cov}  ( v, u ) \; .
\end{eqnarray*}
\end{proposition}
\AfterTheoremEnvironment

Suppose 
Assumptions \ref{GaussianFunctionAssumption} and
\ref{GeneralMeasurementAssumption} hold with
\( V_i (r) = r^2 \) and $K_i$ as given in
\eqref{RepresenterEquation} for \( i = 1 , \ldots , N \).
It follows that \( y \) is Gaussian.
Applying Proposition \ref{ConditionalProposition} with
$u = F(x)$ and $v = y$, we obtain
\( \B{E} (u) = 0 \), \( \B{E} (v) = 0 \), and
\[
\B{E} [ F(x) | y ]
=
\lambda [K_1 (x) \quad  \ldots \quad K_N (x)] 
\left( \lambda \overline{K} + \sigma^2 \B{I}_N \right)^{-1} y \; .
\]
Using the notation $\gamma = \sigma^2 / \lambda$, one obtains
\begin{eqnarray*}
\B{E} [ F(x) | y ]
& = &  
[K_1 (x) \quad  \ldots \quad K_N (x)] 
\left( \overline{K} + \gamma\B{I}_N \right)^{-1} y \; ,
\\ 
& = &  
\sum_{i=1}^N \hat{c}_i K_i (x) \; .
\end{eqnarray*}
where $\hat{c}$ is computed using \eqref{ComputeChat}. 
This shows that in the Gaussian case
the minimum variance estimate
coincides with $\hat{F}$ defined by
\eqref{DefineFhat}.
We formalize this result in the following proposition.

\begin{proposition}
\label{GaussianMinimumVarianceProposition}
Suppose that $F$ satisfies Assumption~\ref{GaussianFunctionAssumption}
and $\B{p}(y|F)$ satisfies Assumption~\ref{GeneralMeasurementAssumption}
with $V_i (r) = r^2$.
Then the minimum variance estimate of $F(x)$ given $y$ is 
$\hat{F}(x)$ defined by \eqref{DefineFhat},
with $ \gamma = \sigma^2 / \lambda$ and
$\mathcal{H}$ the RKHS induced by $K$.
\end{proposition}
\AfterTheoremEnvironment

\subsection{Non-Gaussian measurements: MAP estimate}

We now consider what happens when the Gaussian assumptions on 
$e_i$ are removed. 
If the probability density function 
for $F$ was well defined and given by
\begin{equation*}
\B{p} ( F ) \propto \exp \left( 
	-\frac{\| F \|_{\mathcal{H}}^2 }{2\lambda}
\right) \; ,
\end{equation*}
then the posterior density conditional on the data would be
\begin{equation*}
 \B{p} ( F | y ) 
\propto  
\exp\left(
	- \sum_{i=1}^N \frac{V_i[ y_i - F( x_i ) ]}{2\sigma^2}
	-\frac{\| F \|_{\mathcal{H}}^2 }{2 \lambda}
\right) \; .
\end{equation*}
In this case, the negative log of $\B{p}(F|y)$
would be proportional to the objective
in \eqref{DefineFhat}. 
Hence, one could immediately conclude that $\hat{F}$ is the MAP estimator. 
Unfortunately, the posterior density of $F$ on a function space
is not well defined. 
However, one can consider the MAP estimates
corresponding to any finite sample of $F$
that includes the observations $y_i$
(since these are finite dimensional estimation problems).
The following proposition shows
that $\hat{F}$ solves all such problems.

\begin{proposition} 
\label{GeneralMapProposition}
Suppose that $F$ satisfies Assumption~\ref{GaussianFunctionAssumption}
and $\B{p}(y | F)$ satisfies Assumption~\ref{GeneralMeasurementAssumption}.
Let
$ \{ x_i ~ : ~ i = N+1 , \ldots , N+M \}$ be an arbitrary
set of points in $\mathcal{X}$
where $M$ is a given non-negative integer, and define
\[
f = [ F( x_1 ) , \ldots , F( x_{N+M} ) ]^\R{T} \; .
\]
Then the MAP estimate for $f$ given $y$ is 
\[
\arg \max_f \B{p}( y | f) \B{p} ( f )
=
[ \hat{F}( x_1 ) , \ldots , \hat{F}( x_{N+M} ) ]^\R{T} \; ,
\]
where $\hat{F}$ is defined by \eqref{DefineFhat},
with $ \gamma = \sigma^2 / \lambda$ and
$\mathcal{H}$ is the RKHS induced by $K$.
\end{proposition}
\AfterTheoremEnvironment

\subsection{Non-Gaussian measurements: minimum variance estimate}

When considering non-Gaussian measurement loss functions,
the minimum variance estimate $\B{E}[ F ( \cdot ) | y ]$ and
the MAP estimate $\hat{F} ( \cdot )$ are different.

\begin{example}
Consider the case where $N = 1$, $M = 0$, $V_1 (r) = |r|$, $y = 1$, and
$\lambda = 1$, $\sigma = 1$, $K( x_1 , x_1 ) = 1$.
For this case, $f = F( x_1 )$,
and the MAP estimate for $f$ given $y$ is
\[
\hat{f} = \arg \min_f ( f^2 + | 1 - f | ) = 1 / 2 \; .
\]
Define $A > 0$ by
\[
A 
= 
\int_{-\infty}^{+\infty} \exp ( - f^2 - | 1 -  f | ) \B{d} f \; .
\]
The difference between the minimum variance estimate and the MAP estimate is 
(see Appendix \ref{ExampleMinimumVarianceProof} for details)
\begin{equation}
\label{ExampleMinimumVarianceEstimate}
\B{E} ( f | y ) - \hat{f}
= 
\frac{\exp(-3/4)}{A} \int_{1/2}^{+\infty} 
s \, \frac{ \exp ( 1 - 2 s ) - 1 }{ \exp( s^2 ) } \B{d} s \; .
\end{equation}
For $s > 1/2$, the integrand in \eqref{ExampleMinimumVarianceEstimate}
is negative, so
the right hand side is negative, and $\B{E} ( f | y ) < \hat{f}$.
\end{example}

The following proposition shows that
the minimum variance estimate $\B{E}[ F ( \cdot ) | y]$ and
the MAP estimate $\hat{F}(\cdot)$
belong to the same subspace of $\mathcal{H}$, namely, the
linear span of the functions $K_i (\cdot) , \; i = 1 , \ldots , N$.

\begin{proposition} 
\label{GeneralMinimumVarianceProposition}
Suppose that $F$ satisfies Assumption~\ref{GaussianFunctionAssumption}
and $\B{p}(y | F)$ satisfies Assumption~\ref{GeneralMeasurementAssumption}.
Define
\begin{eqnarray*}
g       & = & [ F( x_1 ) , \ldots , F( x_N ) ]^\R{T} \; ,
\\
\hat{d} & = & \overline{K}^{-1} \B{E} ( g | y ) \; .
\end{eqnarray*}
For each $x \in \mathcal{X}$ the minimum variance estimate of $F(x)$ is
\begin{equation}
\label{GeneralMinimumVarianceEstimate}
\B{E}[ F(x) | y ] = \sum_{i=1}^N \hat{d}_i K_i(x) \; .
\end{equation}
\end{proposition}
\AfterTheoremEnvironment

Note that, given $\sigma$ and $\lambda$,
the vector $\B{E} ( g | y )$ can be approximated using the relation 
\begin{equation*}
\B{p} ( g | y ) 
\propto  
\exp\left(
	- \sum_{i=1}^N \frac{V_i[ y_i - g_i ) ]}{2\sigma^2}
	- \frac{g^T \overline{K}^{-1} g}{2 \lambda}
\right) 
\end{equation*}
together with random sampling technique such as MCMC.

\section{Function and regularization parameter estimation}
\label{RegularizationParameterSection}

In real applications, the regularization parameter
$\gamma = \sigma^2 / \lambda$
is typically unknown and needs to be inferred from data.
In the case of Gaussian measurement noise, this problem is often solved
by exploiting the stochastic interpretation given by
Proposition \ref{GaussianMinimumVarianceProposition}.
For example, following an empirical Bayes approach,
the marginal likelihood can be computed analytically and
the unknown parameters (often called {\it hyperparameters})
can be estimated by optimizing this likelihood, e.g.
see \cite{MacKayNC92} and \cite[Subsection 5.4.1]{Rasmussen}.
$\gamma$ is then set to its estimated value, and $\hat F$ in~\eqref{DefineFhat}
is obtained using equations \eqref{ComputeChat} and \eqref{RepresenterEquation}.
Propositions
\ref{GeneralMapProposition} and \ref{GeneralMinimumVarianceProposition}
provide the statistical foundations that extend this technique
to non-Gaussian measurement noise.

In the more general case of Assumption~\ref{GeneralMeasurementAssumption}
(non-Gaussian measurement noise)
the marginal likelihood cannot be computed analytically.
Let $\eta$ denote the vector of unknown hyperparameters
($\sigma$ and/or $\lambda$) and
recall the notation $g = [ F( x_1) , \ldots , F( x_N ) ]^\R{T}$.
Following a Bayesian approach, we model $\eta$ as
a random vector with prior probability density $\B{p}( \eta )$.
The conditional density for the data $y$
and the unknown function samples $g$,
given the hyperparameters $\eta$ is
\begin{equation*}
\B{p} ( y , g | \eta )
\propto
\prod_{i=1}^N
\frac{1}{\sigma \sqrt{ \lambda } }
\exp \left(
	- \frac{ V_i ( y_i - g_i ) }{2 \sigma^2}
	- \frac{g^T \overline{K}^{-1} g}{2 \lambda}
\right) \; .
\end{equation*}
The difficulty underlying the estimation of $\eta$ is that
$\B{p}( \eta | y )$ is not, in general, available in closed form.
One possibility is to use stochastic simulation techniques, e.g.
MCMC \cite{Gilks} or particle filters \cite{ParticleMCMC}, 
which can sample from $\B{p}( \eta , g | y)$ provided that a suitable
proposal density for $\eta$ and $g$ can be designed.
An MCMC scheme for sampling from the posterior for
$g$ and $\eta$ (corresponding to the $\ell_1$ measurement model)
is described in Appendix \ref{ell1MCMCAppendix} and applied in
section~\ref{SimulationExampleSection} below.
Proposition \ref{GeneralMinimumVarianceProposition} is especially important
because it shows how to compute $\B{E}[F(x) | y]$ for any $x$ from
the minimum variance estimate for $g$. 
Simillary, given an estimate of $\eta$,
we can use Proposition \ref{GeneralMapProposition}
to compute the corresponding $\hat{F}(x)$ for any $x$.

\section{Simulation example}
\label{SimulationExampleSection}

\begin{figure*}
\begin{center}
	\begin{tabular}{cc}
	\hspace{.1in}
	{ \includegraphics[scale=0.38]{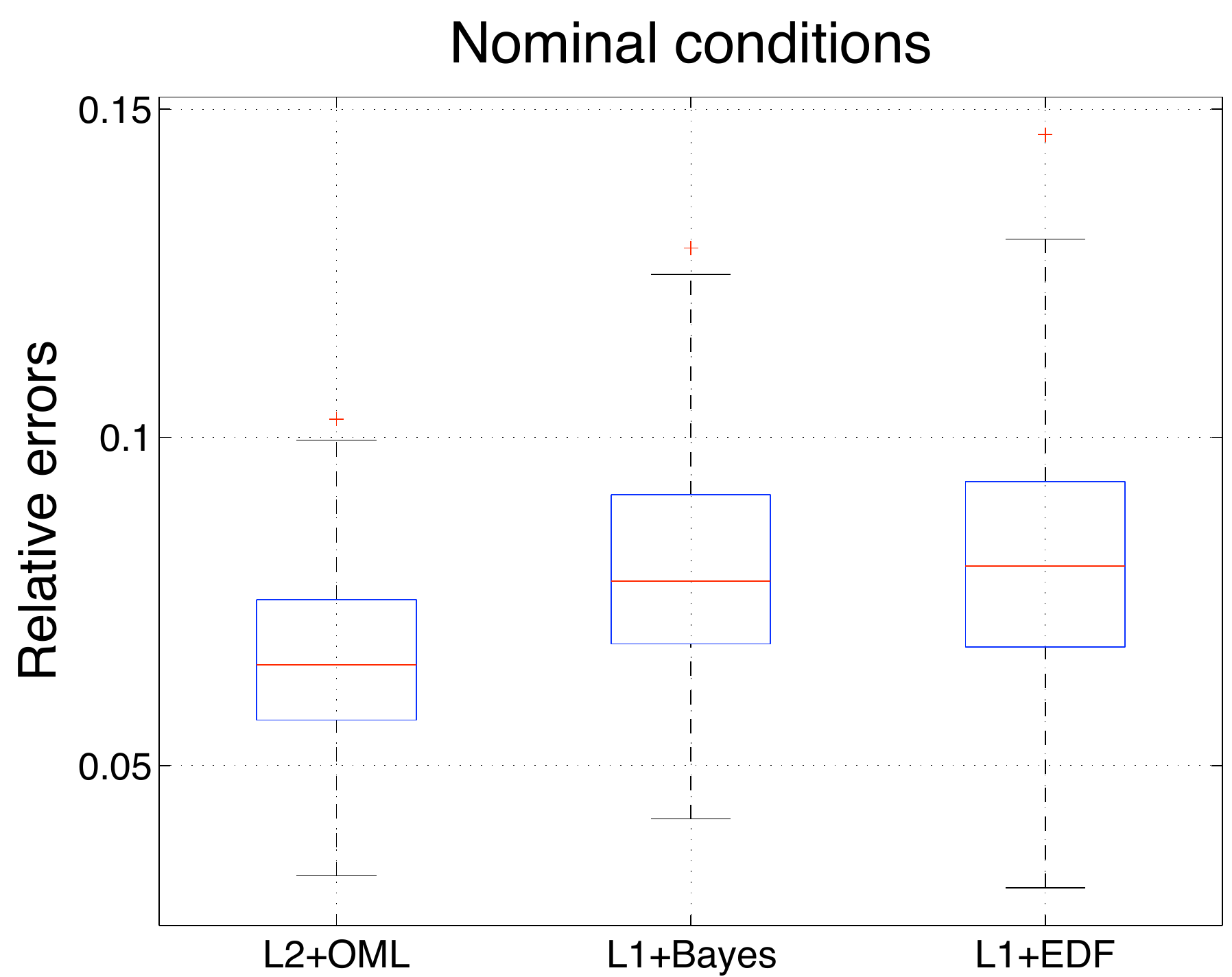}}
	\hspace{.1in}
 	{ \includegraphics[scale=0.38]{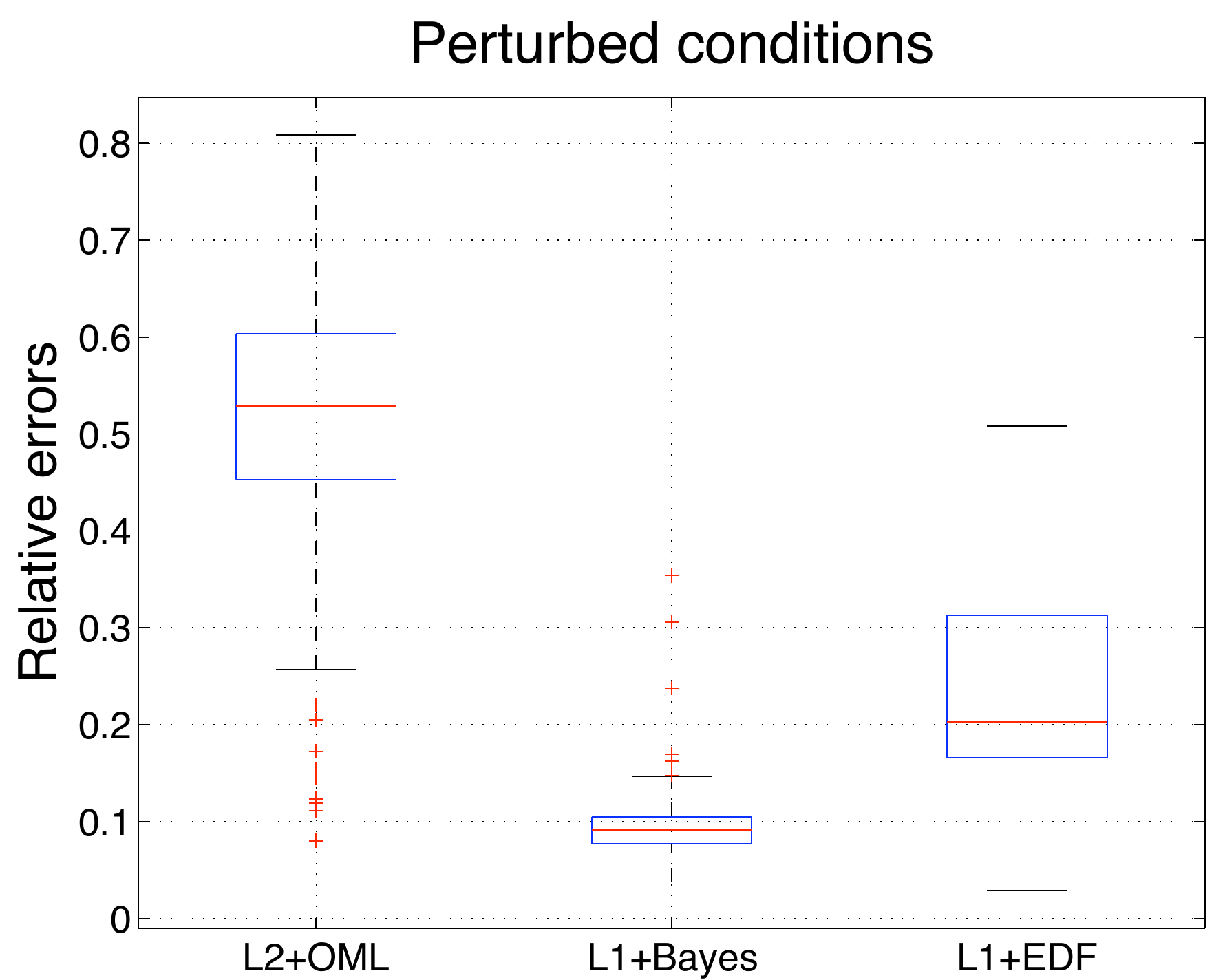}} \\
	\hspace{.1in}
	{\includegraphics[scale=0.42]{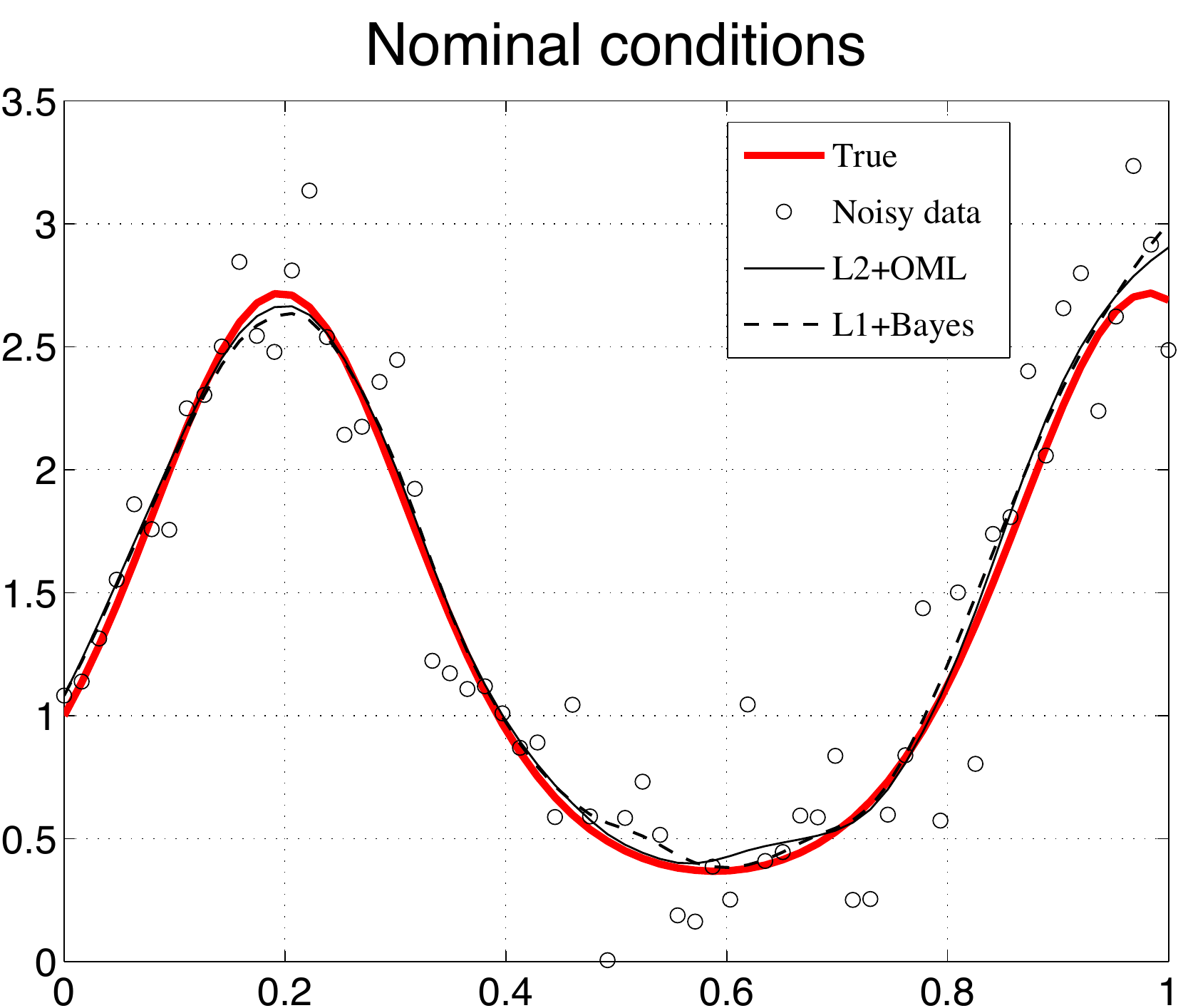}}
	\hspace{.1in}
	{\includegraphics[scale=0.42]{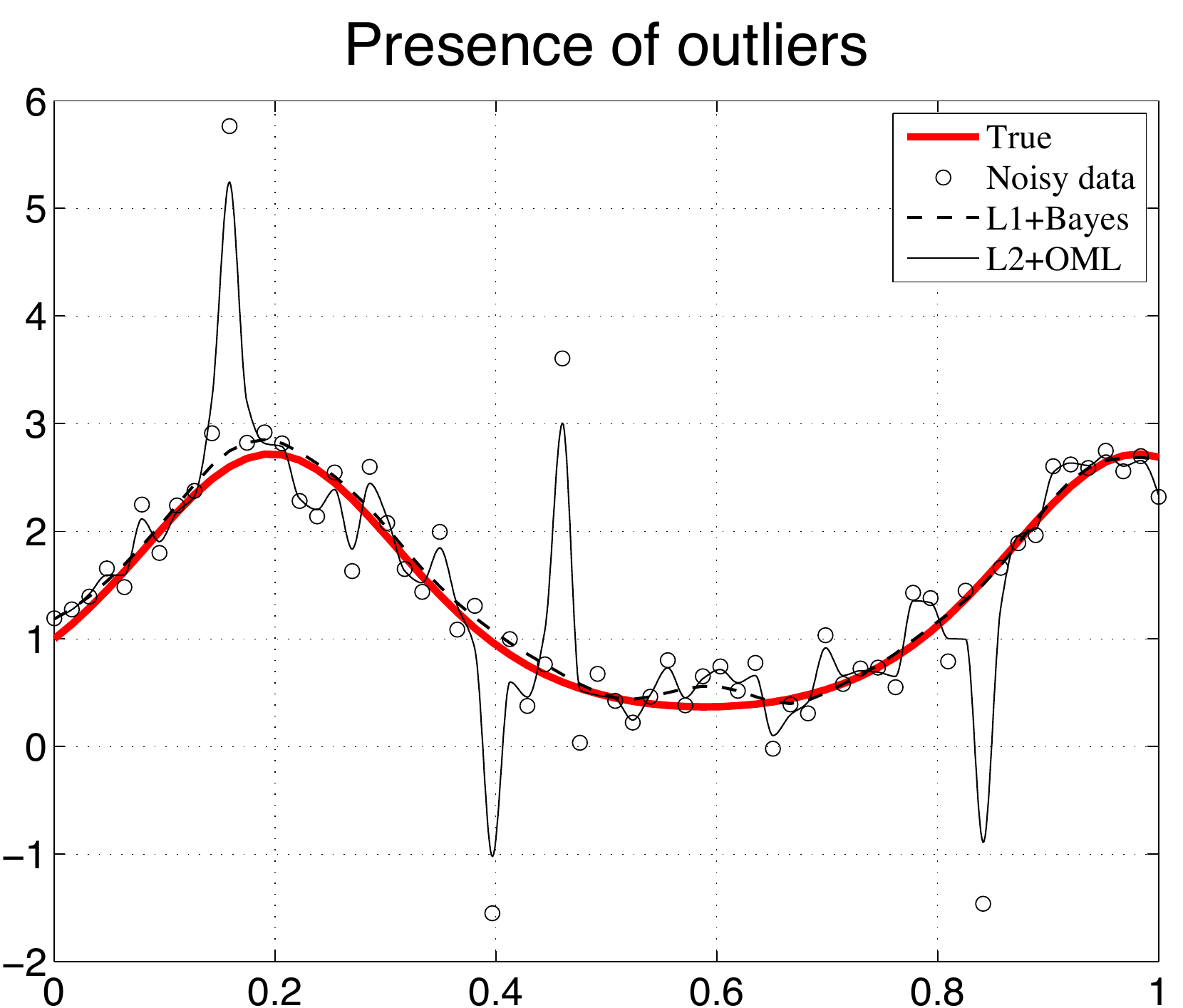}}
	\end{tabular}
    \caption{Simulation. 
		{\it{Top}} 
		Boxplot of the 300 relative errors under nominal (top left) and 
		perturbed (top right) conditions. 
		{\it{Bottom}} True function (thick line), noisy output samples
		($\circ$) and estimates using the $L_2$+OML (solid line) 
		and $L_1$+Bayes (dotted line) 
		estimators under nominal (bottom left) and perturbed (bottom right) 
		conditions.
	} 
	\label{SimulationFigure}
\end{center}
\end{figure*}

We consider the simulated problem in \cite[Section 5.1]{Dinuzzo07}.
The unknown function to be estimated is
$$
F_0(x) = \exp [ \sin(8x) ] \; , \quad 0 \leq x \leq 1
$$
which is displayed as the thick line in
the bottom two panels of Fig. \ref{SimulationFigure}.
This function is reconstructed from the measurements
$$
y_i = F_0(x_i) + e_i
\quad \mbox{with} \quad x_i = (i-1) / 63 \; , \quad i=1,\ldots,64 \; .
$$
We include two Monte Carlo experiments each consisting of 300
function reconstructions.
In the first experiment, for each reconstruction,
measurements $y_i$ are generated using $e_i \sim \B{N} (0, 0.09)$.
A typical data set is plotted as circles $\circ$
in the bottom left panel of Fig. \ref{SimulationFigure}.
In the second experiment, we simulate the presence of outliers
by adding, with probability 0.1, a random offset equal to $\pm 3$ 
to each measurement generated in the first experiment.
A typical data set is plotted as circles in the bottom right panel
of Fig. \ref{SimulationFigure}.

Both experiments compare three different methods for 
modeling the measurement noise and estimating
the kernel scale factor $\lambda$ (described below).
All the methods model the function correlations by using a cubic spline kernel
shifted by 1 to deal with non null initial conditions of $f$ at 0, i.e. $K( x_i , x_j )$ equals
\[
( x_i + 1)  (x_j+1)  \frac{\min( x_i+1 , x_j+1 )}{2} - \frac{\min( x_i+1 , x_j+1 )^3}{6}
\]
\cite[Chapter 1]{Wahba1990}. 
In addition, once an estimate for $\lambda$ is determined,
all methods use the MAP estimator \eqref{DefineFhat}
to reconstruct the function $F_0 (x)$ by solving the problem
in equation \eqref{DefineChat}.

\begin{itemize}

\item
\textit{$L_2$+OML}:
The measurement noise is modeled by a quadratic loss with $\sigma^2 = 0.09$.
(During the second experiment,
the outliers represent unexpected model deviations.)
For each reconstruction, the kernel scale factor $\lambda$
is estimated using marginal likelihood optimization
\cite[section 5.4.1]{Rasmussen}.

\item
\textit{$L_1$+Bayes}.
The measurement noise is modeled by the $\ell_1$ loss with $\sigma$
chosen so the variance of the corresponding Laplace distribution is 0.09.
The kernel scale factor $\lambda$
is estimated by following the Bayesian approach discussed 
(for non-Gaussian noise) in 
section \ref{RegularizationParameterSection}.
More details can be found in Appendix \ref{ell1MCMCAppendix}.
Once the estimate for $ \lambda $ is determined,
the problem in \eqref{DefineChat} is solved using the interior point method 
described in \cite{AravkinCDC12}.

\item
\textit{$L_1$+EDF}.  
The measurement noise is modeled by the $\ell_1$ loss with $\sigma$
chosen so the variance of the corresponding Laplace distribution is 0.09.
The kernel scale factor $\lambda$
is estimated using the approach
described in \cite{Dinuzzo07}; i.e.,
relying on $C_p$-like statistic and the concept 
of equivalent degrees of freedom (EDF). 
The notation $C$ in \cite[eq. 1]{Dinuzzo07}, 
corresponds to $\sigma^{-2} / 2$ in this paper.
The objective in \cite[eq. 19]{Dinuzzo07}  
is optimized on a grid containing 50 values of $\log_{10}(C)$  
uniformly distributed on $[1,6]$. 
The number of degrees of freedom entering \cite[eq. 19]{Dinuzzo07}, 
as a function of $C$, is determined
at every run as described in \cite[Remark 1]{Dinuzzo07} 
(with $\epsilon=0$).
\end{itemize}

The top panels of Fig. \ref{SimulationFigure} are boxplots of the 300 
relative errors defined by
\[
\sqrt{
	\frac{
		\sum_{i=1}^{64} [ F_0(x_i)-\hat{F}(x_i) ]^2
	}{
		\sum_{i=1}^{64} F_0^2(x_i)
	}
}
\]
for the three different methods.
In absence of outliers (top left panel), 
all the methods provide accurate function reconstructions, 
and the $L_2$+OML method performs best. 
The bottom left panel contains the 
results of a single reconstruction using the $L_2$+OML and $L_1$+Bayes methods.

The situation dramatically changes in presence of outliers (top right panel).
As expected, the errors for the $L_2$+OML method increase significantly. 
The estimate obtained by the $L_2$+OML method for a single reconstruction 
is displayed in the bottom right panel (solid line).
It is apparent that the quadratic loss is very vulnerable to 
unexpected model deviations. 
On the other hand the estimate obtained by $L_1$+Bayes method
is much closer to the truth. 
This remarkable performance is confirmed by the top right panel.
The errors corresponding to the $L_1$+Bayes method with outliers
is similar to the performance obtained in the absence of outliers. 
In addition, the $L_1$+Bayes method outperforms the $L_1$+EDF method.  

\begin{remark}
The MCMC scheme 
discussed in the last part of Appendix \ref{ell1MCMCAppendix} was also used
to compute the minimum variance estimate of $F$. 
The performance of this estimator is virtually identical to that of 
$L_1$+Bayes.
Once the MCMC samples are computed, 
there is very little extra computation required to obtain
the minimum variance estimate of $F$.
In addition, it does not require the 
somewhat complex optimization procedure described in \cite{AravkinCDC12}.
\end{remark}

\begin{remark}
We also considered a third experiment,
where the true value of the noise variance, i.e. $\sigma^2=0.99$,
is provided to the three estimators. 
The average error of the $L_2$+OML method decreases from 0.52 to 0.21,
while that of the $L_1$+EDF method decreases from 0.24 to 0.15.
The average error of the $L_1$+Bayes method 
does not change significantly,
staying around $0.1$ in both the second and third experiments. 
\end{remark}

\section{Conclusion}
\label{ConclusionSection}

When the RKHS induced by $K$ is infinite-dimensional,
the realizations of the Gaussian random field with autocovariance $K$
do not fall in $\mathcal{H}$ with probability one, see
\cite[eq. 34]{Parzen63} and also 
\cite{Kallianpur70,Driscoll,Lukic} for generalizations. 
A simple heuristic argument illustrating this fact can be also found in
Chapter 1 of \cite{Wahba1990}. 
The intuition here is that the realizations of $F$ are much less regular than
functions in the RKHS whose kernel is equal to the autocovariance $K$.
On the other hand, 
in the case of Gaussian measurement noise,
$\hat{F}$ defined in \eqref{DefineFhat} is the minimum variance estimate; see
Proposition \ref{GaussianMinimumVarianceProposition}.
In this note we proved a formal connection between Bayesian estimation
and the more general case
prescribed by Assumption~\ref{GeneralMeasurementAssumption}.
Given the training set \( \{ ( x_i , y_i ) \} \),
for any finite set of locations which include the training locations
$\{ x_i \}$, the MAP estimate of $F$ at the locations is
the RKHS estimate evaluated at these locations.
We have also shown that the MAP estimate of $F$
and the minimum variance estimate of $F$
belong to the finite dimensional subspace (of the RKHS)
induced by the covariance $K$ at the training locations.
(These results can be extended to more general cases by using more
general versions of the Representer Theorem \eqref{RepresenterEquation}.)
This link between statistical estimation and RKHS regularization
provides a foundation for the application of statistical approaches
to joint estimation of the function and the regularization parameters.
The simulation example in this paper
illustrates the utility of this connection.

\section{Appendix}
\label{AppendixSection}

\subsection{Lemmas}
We begin the appendix 
with two lemmas which are instrumental in proving
Proposition \ref{GeneralMapProposition}:

\begin{lemma}
\label{GaussianConditionalMaximumLemma} 
Suppose that \( g \) and \( h \) are
jointly Gaussian random vectors. It follows that 
\begin{align*}
& \max_h \log \B{p} ( h | g ) = 
\\
& - \log \det 
\left\{ 2\pi
\left[ \R{cov}( h, h ) - \R{cov} ( h, g ) \R{cov} ( g, g )^{-1} \R{cov} ( g, h ) \right]
\right\}  / 2
\; ,
\end{align*}
and this maximum does not depend on the value of \( g \).
\end{lemma}

\begin{IEEEproof} 
The proof comes from well known properties of joint Gaussian vectors, see e.g.
\cite{Anderson:1979}. 
The conditional density \( \B{p} ( h | g ) \) is Gaussian and is given by
\begin{align*}
- 2 \log \B{p} ( h | g )  
& = \log \det [ 2 \pi \R{cov} ( h, h | g ) ]
\\
& + 
[ h - \B{E} ( h | g ) ]^\R{T} 
	\R{cov} ( h, h | g )^{-1} 
		[ h - \B{E} ( h | g ) ] \; ,
\end{align*}
where, recalling also Proposition \ref{ConditionalProposition}, 
\[
\R{cov} ( h, h | g )
=
\R{cov}( h, h ) - \R{cov} ( h, g ) \R{cov} ( g, g )^{-1} \R{cov} ( g, h ) 
\; .
\]
Thus, \( \R{cov} ( h, h | g ) \) does not depend on the value of \( g \)
(and it would not make sense for it to depend on the value of $h$).
Hence, one has
\begin{align*}
\arg \max_h \B{p} (h | g) & = \B{E} (h | g) \; ,
\\
\max_h \log \B{p} ( h | g )  
& =
- \log \det [ 2 \pi \R{cov} ( h, h | g ) ] / 2
\; .
\end{align*}
This equation, and the representation for $\R{cov} ( h, h | g )$ above
completes the proof of this lemma.
\end{IEEEproof}

\begin{lemma}
\label{MarkovSeparationLemma}
Assume that \( g \) and \( h \) are jointly Gaussian random
vectors and that \( y \) is a random vector such that
\(  \B{p} ( y | g, h ) = \B{p} ( y | g ) \), and
suppose we are given a value for \( y \).
Define the corresponding
estimates for \( g \) and \( h \) by
\[
( \hat{g} , \hat{h} )
=
\arg \max_{g, h}  \B{p} ( y, g , h)
\; ,
\]
and assume the above maximizers are unique. It follows that
\begin{eqnarray}
\label{GaussianSeparationGhatEquation}
\hat{g}
& = &
\arg \max_{g}  \B{p} (y | g )  \B{p} (g) \; ,
\\ 
\label{GaussianSeparationHhatEquation}
\hat{h}
& = &
\arg \max_h  \B{p} ( h |  g = \hat{g} ) \; .
\end{eqnarray}
\end{lemma}

\begin{IEEEproof} We have
\begin{eqnarray*}
\B{p} (y, g, h) 
& = & 
\B{p} (y | g, h )  \; \B{p} ( h | g) \; \B{p} (g) \; ,
\\
& = & 
\B{p} (y | g )  \; \B{p} (g) \; \B{p} ( h | g) \; ,
\\
\max_{g, h} \B{p} (y, g, h) 
& = &
\max_g \left\{ [ \B{p} (y | g )  \; \B{p} (g) ] \; 
	\max_h \B{p} ( h | g) 
\right\} \; .
\end{eqnarray*}
It follows from 
Lemma \ref{GaussianConditionalMaximumLemma} that 
$\max_h \B{p} ( h | g )$ is constant with respect to $g$.
Hence
\(
\hat{g} = \arg \max_g [ \B{p} (y | g )  \; \B{p} (g) ] \; ,
\)
which completes the proof of \eqref{GaussianSeparationGhatEquation},
and
\[
\max_{g, h} \B{p} (y, g, h) 
=
\B{p} (y | \hat{g} )  \; \B{p} ( \hat{g} ) \max_h \B{p} ( h | g = \hat{g} )
\; ,
\]
which completes the proof of \eqref{GaussianSeparationHhatEquation}.
\end{IEEEproof}

\subsection{Proof of Proposition \ref{GeneralMapProposition}}

The kernel matrix $\overline{K}$ is positive definite
and hence invertible (Assumption~\ref{GaussianFunctionAssumption}).
Define the random vectors $g$ and $h$  by
\begin{eqnarray*}
g & = & [ F( x_1 ) , \ldots , F( x_N ) ]^\R{T} \; ,
\\
h & = & [ F( x_{N+1} ) , \ldots , F( x_{N+M} ) ]^\R{T} \; .
\end{eqnarray*}
It follows that $f$ in Proposition~\ref{GeneralMapProposition} 
is given by $f = ( g^\R{T} ,  h^\R{T} )^\R{T}$. 
Notice that $ \B{p} (y | f ) = \B{p} ( y | g ) $  and that
Lemma \ref{MarkovSeparationLemma} can be applied.
From  \eqref{GaussianSeparationGhatEquation} and the hypotheses above,
we obtain
\begin{eqnarray*}
\hat{g}
& = &
\arg \max_{g}  \B{p} (y | g )  \B{p} (g) \; ,
\\
& = &  
\arg \max_{g} \left(
	\frac{1}{2 \sigma^2}
	\sum_{i=1}^N
		V_i [ y_i - g_i ]
		+  
		\frac{g^\R{T} \overline{K}^{-1} g}{2\lambda} 
\right) \; ,
\end{eqnarray*}
Using the representation $g = \overline{K} c$ we obtain
\begin{align*}
\hat{c}
& = &
\arg \max_{c} \left(
	\frac{1}{2 \sigma^2}
	\sum_{i=1}^N 
		V_i \left[ y_i - \sum_{j=1}^N K( x_i , x_j) c_j \right]
		+  
		\frac{c^\R{T} \overline{K} c}{2\lambda} 
\right) \; .
\end{align*}
This agrees with \eqref{DefineChat},
because $\gamma = \sigma^2 / \lambda$, and
thereby shows 
\[
	\hat{g} = [ \hat{F} ( x_1 ) , \cdots , \hat{F} ( x_N ) ]^\R{T} \; .
\]
Finally, by 
Proposition \ref{ConditionalProposition} and 
Lemma \ref{GaussianConditionalMaximumLemma} in conjuction with
\eqref{RepresenterEquation},
\eqref{GaussianSeparationHhatEquation},
and the expression for $\hat{g}$ above,
we obtain
\begin{eqnarray*}
\hat{h} 
& = &
\R{cov}  ( h, g )  \R{cov} ( g, g )^{-1} \hat{g}
\\
& = &
\R{cov}  ( h, g )  ( \lambda \overline{K} )^{-1} ( \overline{K} \hat{c} )
\; ,
\\
& = & 
\left(
\begin{tabular}{ccc}
$K_1 (x_{N+1})$  & \ldots   & $K_N (x_{N+1})$ \\
\vdots           & $\ddots$ & \vdots          \\
$K_1 (x_{N+M})$  & \ldots   & $K_N (x_{N+M})$
\end{tabular}
\right)
\left(
\begin{tabular}{c}
$\hat{c}_1$ \\
$\vdots$\\
$\hat{c}_N$
\end{tabular}
\right)
\; ,
\\
& = &
[ \hat{F} ( x_{N+1} ) , \ldots , \hat{F} ( x_{N+M} ) ]^\R{T}
\; .
\end{eqnarray*}
Combining this with the formula for $\hat{g}$ above,
we conclude
\[
[ \hat{F} ( x_1 ) , \ldots , \hat{F} ( x_{N+M} ) ]^\R{T}
=
\arg \max_f \B{p}(y, f)
\; ,
\]
which completes the proof of Proposition~\ref{GeneralMapProposition}.

\subsection{Proof of Proposition \ref{GeneralMinimumVarianceProposition}}
To obtain the representation (\ref{GeneralMinimumVarianceEstimate})
we compute $\B{E}[F(x) | y]$ by first projecting $F(x)$
onto $g$ and then onto $y$, i.e. using the equivalence
\[
\B{E}[ F(x) | y ]  = \B{E}( \; \B{E}[F(x) | g] \; | y) \; .
\]
Exploiting Proposition \ref{ConditionalProposition}, 
and recalling that $ \R{cov}( g, g ) = \overline{K}$, 
the first projection is given by
$$
\B{E}[F(x) | g] 
=  \R{cov}[ F(x), g ] \R{cov}( g, g )^{-1} g
=  a^\R{T} \overline{K}^{-1} g
$$
where $a \in \B{R}^N$ and $a_i = \R{cov} [ F(x), g_i ] = K_i (x)$.
The second projection yields
\[
\B{E}( \; \B{E}[F(x) | g] \; | y)
=
a^\R{T} \overline{K}^{-1} \B{E}( g | y)
=
\sum_{i=1}^N \hat{d}_i K_i (x)
\]
where $\hat{d}=  \overline{K}^{-1} \B{E}( g | y)$,
which completes the proof.

\subsection{Proof of eq. \eqref{ExampleMinimumVarianceEstimate}}
\label{ExampleMinimumVarianceProof}
It follows from $N = 1$, $\gamma = 1$, 
that $c$ is a scalar, $f = F( x_1 ) = c$, and
using \eqref{DefineChat} we have
\[
\hat{f} = \hat{c} = \arg \min_c | 1 - c | + c^2 = 1/2
\]
It also follows that
\[
\B{p} (y | f) \B{p}(f) \propto \exp ( - f^2 - | 1 - f | )
\]
The minimum variance estimate $\B{E}( f | y )$,
and its difference from the map estimate $\hat{f}$,
are given by
\begin{align*}
\B{E} ( f | y ) 
& = 
\frac{1}{A} \int_{-\infty}^{+\infty} f \exp ( - f^2 - | 1 -  f | ) \B{d} f \; ,
\\
\B{E} ( f | y ) - \hat{f}
& =
\frac{1}{A} \int_{-\infty}^1 (f - 1/2 ) \exp ( - f^2 - 1 +  f ) \B{d} f
\\
& +
\frac{1}{A} \int_1^{+\infty} 
	(f - 1/2 ) \exp ( - f^2 + 1 -  f ) \B{d} f \; .
\end{align*}
Multiplying both sides of the equation by $A$ and
using the change of variables $s = f - 1/2$, we obtain
\begin{align*}
& A ( \B{E} ( f | y ) - \hat{f} )
\\
& = 
\int_{-\infty}^{1/2} s e^{- (s + 1/2)^2 + s - 1/2} \B{d} s
+
\int_{1/2}^{+\infty} 
	s e^{- (s + 1/2)^2  -  s + 1/2 } \B{d} s \; ,
\\
& =
\int_{-\infty}^{1/2} s e^{ - s^2 - 3/4} \B{d} s
+
\int_{1/2}^{+\infty} 
	s e^{-s^2 - 2 s  + 1/4} \B{d} s \; ,
\\
& =
\int_{-\infty}^{-1/2} s e^{ - s^2 - 3/4 } \B{d} s
+
\int_{1/2}^{+\infty} 
	s e^{ -s^2 - 2 s  + 1/4} \B{d} s \; ,
\\
& = 
\int_{1/2}^{+\infty} 
s e^{ - s^2 - 3/4 } \left[ e^{ 1 - 2 s } - 1 \right] \B{d} s \; .
\end{align*}
This completes the proof of \eqref{ExampleMinimumVarianceEstimate}.

\subsection{Details of the MCMC scheme for $L_1$+Bayes}
\label{ell1MCMCAppendix}

If Assumptions 
\ref{GaussianFunctionAssumption} and \ref{GeneralMeasurementAssumption}
hold with $V_i(r) = 2 \sqrt{2} \sigma |r|$,
the noise $e_i$ is Laplacian with variance $\sigma^2$.
In this case,
it can be difficult to build an efficient MCMC scheme to sample 
from the posterior of $\eta$ and $g$.
This is because, a posteriori, the components of 
$g$ are generally strongly correlated.
It is useful to use to a scale mixture of
normals representation because for each normal,
the posterior distribution can be represented in closed form.
To be specific,
each $\B{p} (e_i)$ admits the representation 
\cite{Andrews}
\begin{eqnarray*}
\B{p} (e_i) 
& = & 
\frac{1}{\sqrt{2} \sigma} 
	\exp \left( -\sqrt{2} \, | e_i | \, / \, \sigma \right)
\\
& = &
\int_{0}^{+\infty} 
\frac{1}{\sqrt{2 \pi \tau_i}} \exp \left( - \frac{e_i^2}{2 \tau_i} \right)
	\frac{1}{\sigma^2} \exp \left( -\frac{\tau_i}{\sigma^2} \right )
		\B{d} \tau_i \;.
\end{eqnarray*}
Hence, we can model Laplacian noise $e_i$ as a mixture of 
Gaussians with variances $\tau_i$ that are 
exponential random variables of probability density
\begin{eqnarray}
\B{p} ( \tau_i ) & = &\left\{ \begin{array}{ll}
	\exp \left( - \tau_i \, / \, \sigma^2 \right) \, / \sigma^2
	& \R{if} \; \tau_i \geq 0
	\\
	0 & \R{otherwise}
\end{array} \right.
\nonumber
\\
\B{p} ( \tau  ) & = & \B{p} ( \tau_1 ) \cdots \B{p} ( \tau_N )
\label{TauDensity}
\end{eqnarray}
We restrict our attention to the case where $\eta = \lambda$,
and use $\tau = ( \tau_1 , \ldots , \tau_N )^\R{T}$ 
to denote the independent random variables
(which are also indepdent of $\lambda$).
We have
$$
\B{p} ( \tau , \lambda | y) 
\propto 
\B{p} (y | \tau, \lambda )  \B{p} (\tau)  \B{p} (\lambda)
$$
Given $\tau$ and $\lambda$,
we have the linear Gaussian model $y = g + \xi$, 
where $g$ and $\xi$ are independently distributed according to 
\[
g \sim \B{N}( 0, \lambda \overline{K}) \quad \R{and} \quad
\xi \sim \B{N}[0, \R{diag}( \tau ) ],
\]
where $\R{diag}( \tau )$ is the diagonal matrix with $ \tau $
along its diagonal.
Notice that $\B{p} (y | \tau , \lambda) $
can be computed in closed from using 
the classical Gaussian marginal likelihood result;
e.g., see \cite[subsection 5.4.1]{Rasmussen}.
To be specific, using the notation 
$C( \tau , \lambda )  = \lambda \overline{K} + \R{diag}(\tau)$,
\begin{equation}
\label{YGivenTauLambdaDensity}
\B{p} (y | \tau, \lambda) 
=   
\frac{1}{\sqrt{2 \pi \det[ C( \tau , \lambda ) ]}} 
\exp \left[ - \frac{1}{2} y^\R{T} C ( \tau , \lambda )^{-1} y \right]  \; .
\end{equation}
Using an improper flat prior on $\lambda \ge 0$, we obtain
$$
\B{p} ( \tau , \lambda | y )
\propto \left\{ \begin{array}{ll}
	\B{p} ( y | \tau , \lambda ) \B{p} ( \tau ) 
	& \R{if} \; \lambda \ge 0
	\\
	0 & \R{otherwise}
\end{array} \right.
$$
were $\B{p} ( y | \tau , \lambda )$ can be computed using 
\eqref{YGivenTauLambdaDensity} 
and $\B{p} ( \tau )$ can be computed using 
\eqref{TauDensity}.
We are now in a position to describe the 
MCMC scheme used for the $L_1$+Bayes method 
in Section \ref{SimulationExampleSection}.
The scale factor $\lambda$, and all the components of $\tau$ 
are simultaneously updated using a random walk Metropolis scheme \cite{Gilks}.
The proposal density is independent normal increments with standard deviation 
$30$ and $\sigma^2 / 30$ for $\lambda$ and $\tau_i$ respectively. 
This simple scheme has always led to an acceptance rate over $20\%$.
We have assessed that this follows from the fact that the components
of $\lambda$ and $\tau$ have low correlation a posteriori. 
For each function reconstruction, 
$L = 10^6$ MCMC realizations from $\B{p} ( \tau , \lambda | y)$ 
were obtained by the MCMC scheme
(which we denote by \( \{ \tau^\ell , \lambda^\ell \} \) below).
Using the convergence diagnostics described in \cite{Raftery},
this allowed us to estimate the quantiles
$q= {0.025, 0.25, 0.5, 0.75,0.975}$ of the marginal posterior of 
$\lambda$ with precision
$r= {0.02, 0.05, 0.01, 0.05,0.02}$, respectively, with probability
$0.95$. 

Now consider recovering the minimum variance estimate $\B{E}[ F(x) | y ]$.
We have seen from Proposition \ref{GeneralMinimumVarianceProposition}
that this reduces to computing $\B{E}( g | y)$.
Note that,
given a value for $\lambda$ and $\tau$, $g$ and $y$ are jointly Gaussian.
Applying Proposition \ref{ConditionalProposition}
\begin{eqnarray*}
\B{E} ( g | y : \tau , \lambda )
& = &
\R{cov}( g , y | \tau , \lambda ) \R{cov}(y, y | \tau , \lambda)^{-1} y \; ,
\\
& = &
\lambda \overline{K} C( \tau , \lambda )^{-1} y \; .
\end{eqnarray*}
Hence, it follows that $\B{E}( g | y)$ can be approximated as
$$
\B{E}( g | y ) 
\approx 
\overline{K} 
	\; \frac{1}{L}
		\sum_{\ell=1}^{L} 
			\lambda^\ell C ( \tau^\ell , \lambda^\ell )^{-1} y \; ,
$$
where and $\{ \tau^\ell , \lambda^\ell \}_{\ell=1}^L$ are the realizations 
from $\B{p} ( \tau , \lambda |y)$ 
achieved by the MCMC scheme above described above.

\bibliographystyle{IEEEtran}
\bibliography{rkhs_bayes}

\begin{thebibliography}{10}
\providecommand{\url}[1]{#1}
\csname url@samestyle\endcsname
\providecommand{\newblock}{\relax}
\providecommand{\bibinfo}[2]{#2}
\providecommand{\BIBentrySTDinterwordspacing}{\spaceskip=0pt\relax}
\providecommand{\BIBentryALTinterwordstretchfactor}{4}
\providecommand{\BIBentryALTinterwordspacing}{\spaceskip=\fontdimen2\font plus
\BIBentryALTinterwordstretchfactor\fontdimen3\font minus
  \fontdimen4\font\relax}
\providecommand{\BIBforeignlanguage}[2]{{%
\expandafter\ifx\csname l@#1\endcsname\relax
\typeout{** WARNING: IEEEtran.bst: No hyphenation pattern has been}%
\typeout{** loaded for the language `#1'. Using the pattern for}%
\typeout{** the default language instead.}%
\else
\language=\csname l@#1\endcsname
\fi
#2}}
\providecommand{\BIBdecl}{\relax}
\BIBdecl

\bibitem{Aronszajn}
N.~Aronszajn, ``Theory of reproducing kernels,'' \emph{Trans. of the American
  Mathematical Society}, vol.~68, pp. 337--404, 1950.

\bibitem{Scholkopf01b}
B.~Sch\"{o}lkopf and A.~J. Smola, \emph{Learning with Kernels: Support Vector
  Machines, Regularization, Optimization, and Beyond}, ser. (Adaptive
  Computation and Machine Learning).\hskip 1em plus 0.5em minus 0.4em\relax The
  MIT Press, 2001.

\bibitem{Wahba1990}
G.~Wahba, \emph{Spline models for observational data}.\hskip 1em plus 0.5em
  minus 0.4em\relax SIAM, Philadelphia, 1990.

\bibitem{PoggioGirosi}
F.~Girosi, M.~Jones, and T.~Poggio, ``Regularization theory and neural networks
  architecture,'' \emph{Neural Computation}, vol.~7, pp. 219--269, 1995.

\bibitem{Wahba1998}
G.~Wahba, ``Support vector machines, reproducing kernel {H}ilbert spaces and
  randomized {GACV},'' Department of Statistics, University of Wisconsin,
  Technical Report 984, 1998.

\bibitem{Scholkopf01}
B.~Sch\"{o}lkopf, R.~Herbrich, and A.~J. Smola, ``A generalized representer
  theorem,'' \emph{Neural Networks and Computational Learning Theory}, vol.~81,
  pp. 416--426, 2001.

\bibitem{Poggio}
T.~Poggio and F.~Girosi, ``Networks for approximation and learning,''
  \emph{Proceedings of the IEEE}, vol.~78, pp. 1481--1497, 1990.

\bibitem{Anderson:1979}
B.~D.~O. Anderson and J.~B. Moore, \emph{Optimal Filtering}.\hskip 1em plus
  0.5em minus 0.4em\relax Englewood Cliffs, N.J., USA: Prentice-Hall, 1979.

\bibitem{Kimeldorf71Bayes}
G.~Kimeldorf and G.~Wahba, ``A correspondence between {B}ayesan estimation of
  stochastic processes and smoothing by splines,'' \emph{Ann. Math. Statist.},
  vol.~41, no.~2, pp. 495--502, 1971.

\bibitem{Girosi95}
F.~Girosi, M.~Jones, and T.~Poggio., ``Regularization theory and neural
  networks architectures.'' \emph{Neural Computation}, vol.~7, no.~2, pp.
  219--269, 1995.

\bibitem{Rasmussen}
C.~Rasmussen and C.~Williams, \emph{{G}aussian Processes for Machine
  Learning}.\hskip 1em plus 0.5em minus 0.4em\relax The MIT Press, 2006.

\bibitem{Vapnik98}
V.~Vapnik, \emph{Statistical Learning Theory}.\hskip 1em plus 0.5em minus
  0.4em\relax New York, NY, USA: Wiley, 1998.

\bibitem{Evgeniou99}
T.~Evgeniou, M.~Pontil, and T.~Poggio, ``Regularization networks and support
  vector machines,'' \emph{Advances in Computational Mathematics}, vol.~13, pp.
  1--150, 2000.

\bibitem{Gunter06}
L.~Gunter and J.~Zhu, ``Computing the solution path for the regularized support
  vector regression,'' in \emph{Advances in Neural Information Processing
  Systems 18}, Y.~Weiss, B.~Sch\"{o}lkopf, and J.~Platt, Eds.\hskip 1em plus
  0.5em minus 0.4em\relax Cambridge, MA: MIT Press, 2006, pp. 483--490.

\bibitem{Bogachev}
V.~Bogachev, \emph{{G}aussian measures}.\hskip 1em plus 0.5em minus 0.4em\relax
  AMS, 1998.

\bibitem{Gilks}
W.~Gilks, S.~Richardson, and D.~Spiegelhalter, \emph{Markov chain Monte Carlo
  in Practice}.\hskip 1em plus 0.5em minus 0.4em\relax London: Chapman and
  Hall, 1996.

\bibitem{Dinuzzo07}
F.~Dinuzzo, M.~Neve, {G. De Nicolao}, and U.~P. Gianazza, ``On the representer
  theorem and equivalent degrees of freedom of {SVR},'' \emph{J. of Machine
  Learning Research}, vol.~8, pp. 2467--2495, 2007.

\bibitem{NoisePMG2000}
M.~Pontil, S.~Mukherjee, and F.~Girosi, ``On the noise model of support vector
  machine regression,'' in \emph{Proc. of Algorithmic Learning Theory 11th
  International Conference ALT 2000}, Sydney, 2000.

\bibitem{MacKayNC92}
D.~MacKay, ``Bayesian interpolation,'' \emph{Neural Computation}, vol.~4, pp.
  415--447, 1992.

\bibitem{ParticleMCMC}
C.~Andrieu, A.~Doucet, and R.~Holenstein, ``Particle markov chain monte carlo
  methods,'' \emph{Journal of the Royal Statistical Society: Series B
  (Statistical Methodology)}, vol.~72, no.~3, pp. 269--342, 2010.

\bibitem{AravkinCDC12}
A.~Aravkin, J.~Burke, and G.~Pillonetto, ``Nonsmooth regression and state
  estimation using piecewise quadratic log-concave densities,'' in
  \emph{Proceedings of the 51st IEEE Conference on Decision and Control (CDC
  2012)}, 2012.

\bibitem{Parzen63}
E.~Parzen, ``Probability density functionals and reproducing kernel {H}ilbert
  spaces,'' in \emph{Proc. of the Symposium on Time Series Analysis}.\hskip 1em
  plus 0.5em minus 0.4em\relax New York: John Wiley and Sons, 1963.

\bibitem{Kallianpur70}
G.~Kallianpur, ``The role of reproducing kernel {H}ilbert spaces in the study
  of {G}aussian processes,'' in \emph{Advances in Probability and Related
  Topics}.\hskip 1em plus 0.5em minus 0.4em\relax Marcel Dekker, 1970, pp.
  49--83.

\bibitem{Driscoll}
M.~Driscoll, ``The reproducing kernel {H}ilbert space structure of the sample
  paths of a {G}aussian process,'' \emph{Zeitschrift fur
  Wahrscheinlichkeitstheorie und verwandte Gebiete}, vol.~26, pp. 309--316,
  1973.

\bibitem{Lukic}
M.~Lukic and J.~Beder, ``Stochastic processes with sample paths in reproducing
  kernel {H}ilbert spaces,'' \emph{Trans. Amer. Math. Soc.}, vol. 353, pp.
  3945--3969, 2001.

\bibitem{Andrews}
D.~Andrews and C.~Mallows, ``Scale mixtures of normal distributions,''
  \emph{Journal of the Royal Statistical Society, Ser. B}, vol.~36, pp.
  99--102, 1974.

\bibitem{Raftery}
A.~Raftery and S.~Lewis, ``Implementing mcmc,'' in \emph{Markov Chain Monte
  Carlo in Practice}, 1996, pp. 115--130.

\end{thebibliography}
\end{document}